\documentclass{article}

\usepackage{PRIMEarxiv}

\usepackage[utf8]{inputenc}
\usepackage[T1]{fontenc}
\usepackage{hyperref}
\usepackage{url}
\usepackage{booktabs}
\usepackage{amsfonts}
\usepackage{amsmath}
\usepackage{microtype}
\usepackage{graphicx}
\usepackage{booktabs}
\usepackage{graphicx}
\usepackage{amssymb}
\usepackage{pifont}
\usepackage[dvipsnames]{xcolor}
\newcommand{\xmark}{\ding{55}}
\graphicspath{{media/}}

\newcommand{\algname}{\textsc{ArtisanCAD }}
\title{ArtisanCAD: An Industrial-Level CAD Agent with Expert-Grounded Knowledge Distillation}

\author{
\textbf{Yunhan Xu$^{1,\dagger}$ \quad
Qifeng Wu$^{1,\dagger}$ \quad
Xunjin Li$^{4,\dagger}$ \quad
Yuanwei Bin$^{2,4,*}$ \quad
Qingsong Yao$^{3,*}$}\\
\textbf{Jianghang Gu$^{4}$ \quad
Guan Wang$^{4}$ \quad
Weihao Lv$^{4}$ \quad
Huiyu Yang$^{4}$ \quad
Wenfa Luo$^{5,*}$}\\
\textbf{
Jiao Xiang$^{5}$ \quad
Yuntian Chen$^{2}$ \quad
Shiyi Chen$^{2}$}
\\
\\
$^{1}$Peking University, Beijing 100871, China \\
$^{2}$Eastern Institute of Technology, Ningbo 315200, Zhejiang, China \\
$^{3}$Renmin University of China, Beijing 100872, China \\
$^{4}$TenFong Technology Co., Ltd., Shenzhen 518000, Guangdong, China \\
$^{5}$IM Motors Technology Co., Ltd., Shanghai 201210, China \\
\\
$^\dagger$Equal contribution.\\
$^*$Corresponding authors: \texttt{ybin@eitech.edu.cn}, \texttt{qingsongyao@ruc.edu.cn},
\texttt{luowenfa@immotors.com}.
}

\begin{document}
\maketitle

\begin{abstract}
Computer-aided design (CAD) for industrial components requires long-horizon procedural modeling, robust feature dependencies, editable parametric geometry, and production-grade B-Rep execution.
Existing text-to-CAD methods have made promising progress in generating CAD programs from natural-language descriptions, but they still struggle when user prompts are ambiguous, underspecified, or only describe high-level design intent.
They also rarely exploit expert procedural knowledge naturally available in industrial workflows, such as CATIA operation recordings, macro logs, drawing notes, and engineering descriptions.
We present \algname, a skill-guided industrial CAD agent with expert-grounded knowledge distillation.
The core of \algname is CAD intermediate representation (CAD-IR), an executable procedural representation that encodes parameters, ordered operations, MCP tool bindings, dependencies, generated entities, and verification rules.
CAD-IR plays two key roles: it first serves as the carrier for distilling expert CAD procedures into reusable parameterized skills; then it provides a procedural scaffold that turns vague or intermediate-level prompts into complete executable CAD operations.
\algname retrieves expert-derived skills, instantiates and revises CAD-IR, executes the resulting procedure through a dedicated CATIA-MCP backend, and uses multi-view visual feedback for iterative refinement, and finally generates production-ready B-Rep models.
On the Text2CAD benchmark, CAD-IR improves generation from intermediate prompts by reducing mean Chamfer Distance from $14.83$ to $9.88$, showing its ability to bridge ambiguous textual intent and executable CAD construction.
On four complex automotive components, CAD-IR enables expert CATIA recordings to be distilled into reusable skills, allowing \algname to generate editable CATIA-native B-Rep models for new variant requests.
\end{abstract}


\begin{figure}
    \centering
    \includegraphics[width=\textwidth]{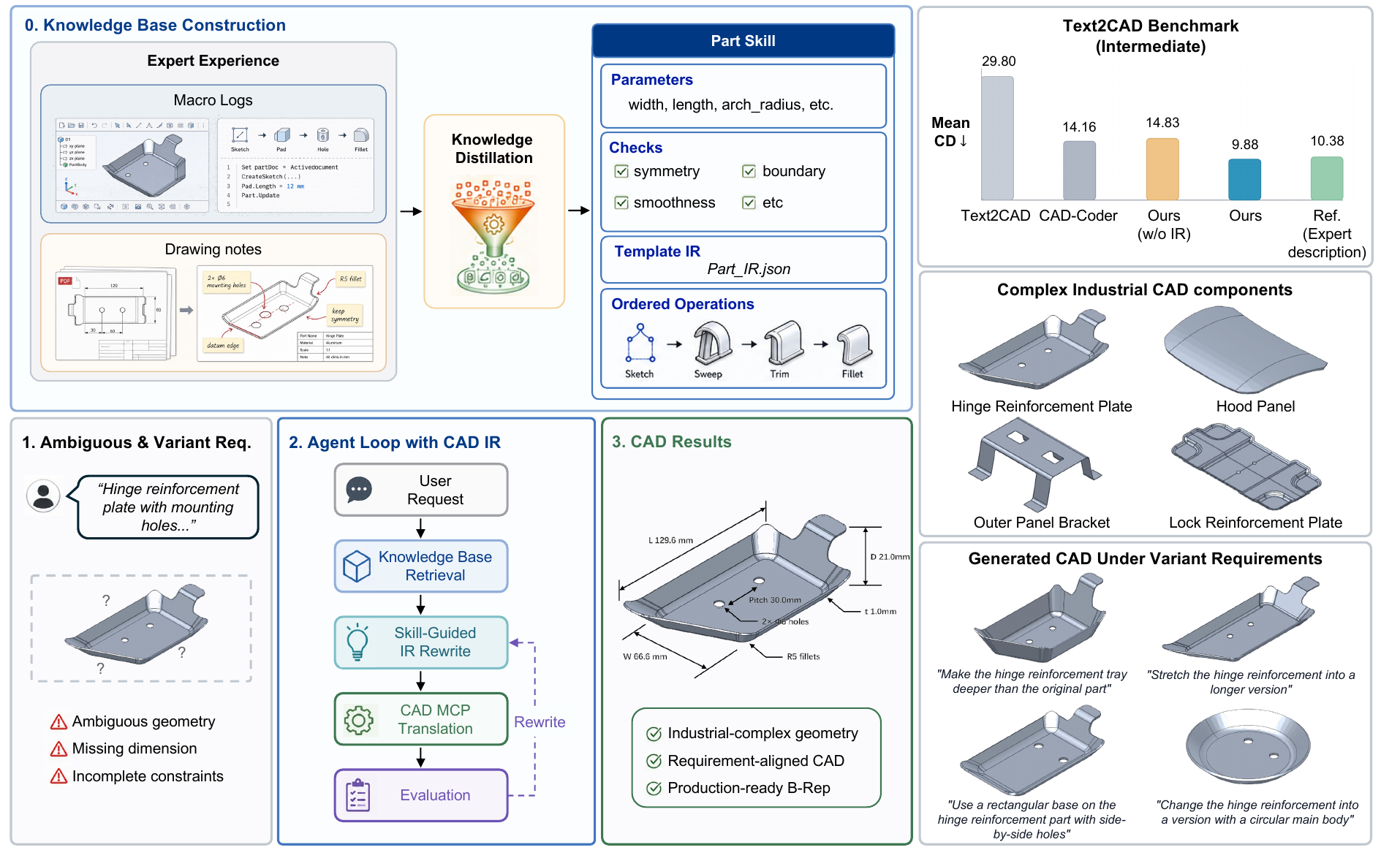}
    \caption{Overview of \algname. Expert CATIA recordings and drawing notes are distilled into reusable part skills with parameters, checks, template CAD-IR, and ordered operations. Given ambiguous variant requests, \algname retrieves the corresponding skill, rewrites CAD-IR through an agent loop, executes it via CAD-MCP, and produces evaluated, editable, production-ready B-Rep CAD models.}
    \label{fig:teaser}
\end{figure}

\section{Introduction}

Computer-aided design (CAD) remains one of the most expertise-intensive workflows in modern engineering and manufacturing.
In industrial sectors such as automotive, aerospace, and consumer mechanical product design, engineers use professional CAD systems such as CATIA, Siemens NX, and Creo to construct editable parametric models with sketches, references, constraints, features, and named entities~\cite{shah1995parametric,camba2016parametric}.
Unlike casual 3D modeling, industrial CAD authoring is rarely a one-shot shape generation problem.
It requires long-horizon procedural modeling, robust feature dependencies, manufacturability-aware construction, and repeated refinement under part-specific requirements such as tolerances, wall thickness, draft constraints, assembly interfaces, and downstream editability.
For complex industrial components, such expertise is accumulated through long-term practice, where engineers learn not only what geometry to create, but also how to organize references, expose parameters, order operations, and avoid fragile modeling procedures.

Recent progress in large language models has motivated a growing body of AI-driven CAD generation methods.
Early CAD generation works formulate CAD as a structured generative language by modeling sketches, constraints, and sketch-and-extrude command sequences~\cite{seff2020sketchgraphs,ganin2021computer,para2021sketchgen,willis2021fusion,wu2021deepcad,xu2022skexgen,li2022free2cad,sharma2018csgnet,jones2020shapeassembly}.
Recent text-to-CAD systems further map natural-language descriptions to parametric CAD programs, command sequences, or executable modeling code~\cite{khan2024text2cad,li2024cad,li2025cad,wang2025cad,guan2026cad,wang2025text,xie2025text}.
More recently, CAD agents have explored tool use, program execution, visual feedback, B-Rep grounding, and iterative correction for more controllable CAD generation~\cite{mallis2025cad,wang2025text,li2026towards}.
These efforts demonstrate that language models can provide an intuitive interface between design intent and executable CAD modeling.

However, a substantial gap remains between existing text-to-CAD systems and industrial CAD practice.
Current methods are still largely developed and evaluated on relatively simple parts or short sketch-extrude programs, while real industrial components require long-horizon procedures with curved B-Rep surfaces, sweeps, splits, datum construction, hybrid surface-solid operations, reusable references, and editable feature dependencies.
In such complex modeling workflows, experienced engineers possess rich procedural knowledge that directly determines the quality, robustness, and efficiency of CAD construction.
They routinely leave behind macro logs, feature trees, operation histories, parameter tables, and modeling templates, which encode how to decompose a component, organize references, parameterize dimensions, order dependent features, and repair modeling failures.
Effectively extracting and reusing this knowledge can benefit industrial CAD generation in two ways: it provides strong procedural priors for producing high-quality, editable, production-oriented CAD models, and it turns repeated expert operations into reusable skills that can improve engineers' productivity~\cite{ni2026trace2skill,xia2026skillrl}.

\begin{table}[t]
\centering
\small
\caption{Comparison with representative accepted text-to-CAD and CAD-agent methods.
Unlike prior methods, \algname explicitly converts expert CAD procedures into executable CAD-IR and reusable skills, and executes them through an industrial CATIA backend.}
\label{tab:method_comparison}
\resizebox{\linewidth}{!}{
\begin{tabular}{lcccccc}
\toprule
Method & Venue & Text Input & Visual Feedback & Executable CAD & Expert Macro-to-Skill & Industrial CATIA Backend \\
\midrule
CAD Translator~\cite{li2024cad} & MM 2024 & \checkmark & \xmark & \checkmark & \xmark & \xmark \\
Text2CAD~\cite{khan2024text2cad} & NeurIPS 2024 & \checkmark & \xmark & \checkmark & \xmark & \xmark \\
CAD-GPT~\cite{wang2025cad} & AAAI 2025 & \checkmark & \xmark & \checkmark & \xmark & \xmark \\
CAD-LLaMA~\cite{li2025cad} & CVPR 2025 & \checkmark & \xmark & \checkmark & \xmark & \xmark \\
CADFusion~\cite{wang2025text} & ICML 2025 & \checkmark & \checkmark & \checkmark & \xmark & \xmark \\
CAD-Assistant~\cite{mallis2025cad} & ICCV 2025 & \checkmark & \checkmark & \checkmark & \xmark & \xmark \\
FutureCAD~\cite{li2026towards} & ICML 2026 & \checkmark & \xmark & \checkmark & \xmark & \xmark \\
\midrule
\algname & Ours & \checkmark & \checkmark & \checkmark & \checkmark & \checkmark \\
\bottomrule
\end{tabular}
}
\end{table}

In this paper, we present \algname, an industrial-level CAD agent that generates complex production-oriented CAD components by grounding CAD generation in expert procedural knowledge.
\algname supports an expert knowledge distillation workflow: expert CATIA operation recordings, macro logs, drawing notes, and engineering descriptions are parsed into structured modeling procedures, converted into CAD intermediate representation (CAD-IR), and further distilled into reusable parameterized skills.
During generation, these skills are retrieved and adapted to user-provided variant requests, enabling the agent to reuse expert operation orders, parameter schemas, and modeling templates instead of rediscovering an industrial CAD workflow from scratch.
\algname is also designed as a multi-backend CAD agent.
Its procedures are represented at the IR level and can be dispatched to different CAD execution backends; in this work, we implement a dedicated CATIA-MCP backend that maps CAD-IR operations to CATIA-native commands and produces editable, production-oriented B-Rep models.
A lightweight vision feedback loop further evaluates multi-view renderings of the generated CAD model and rewrites the CAD-IR when visual or structural inconsistencies are observed.

The core of \algname is \textbf{CAD-IR}, an executable intermediate representation for industrial CAD procedures.
CAD-IR plays two complementary roles.
First, it serves as the carrier for expert knowledge distillation: expert macro recordings are transformed into ordered operations, parameter schemas, backend tool bindings, dependencies, generated entities, and verification rules, which can then be reused as skill templates.
Second, it serves as the procedural scaffold for ambiguous and variant user requests.
Given a request that may only specify high-level changes, CAD-IR allows the agent to fill missing parameters, preserve expert-derived operation order, instantiate the required MCP tool chain, and generate a complete executable CAD procedure that satisfies the intended design constraints.
This design keeps planning, expert knowledge reuse, backend execution, and feedback-based repair within a unified representation.

We evaluate \algname on both public text-to-CAD tasks and industrial automotive CAD cases.
On the Text2CAD dataset, we isolate the effect of CAD-IR by disabling the expert skill library and directly generating CAD-IR from intermediate-level prompts.
Even without component-specific expert knowledge, CAD-IR substantially improves generation quality: compared with direct intermediate-prompt generation, it reduces the mean Chamfer Distance from $14.83$ to $9.88$ and improves mean solid IoU from $0.632$ to $0.654$.
This shows that CAD-IR itself provides a useful procedural scaffold for turning vague textual descriptions into executable CAD operations.
Beyond the public benchmark, we record expert CATIA macro demonstrations for four complex automotive components, distill them into CAD-IR and reusable skills, and ask experts to provide additional variant requests.
The results show that skill-guided CAD-IR generation can produce CATIA-native editable B-Rep models for these industrial components, while generation without expert-derived skills fails to complete the long-horizon modeling procedures.

Our contributions are summarized as follows:
\begin{itemize}
\item We propose an \textbf{expert-to-skill acquisition mechanism} for industrial CAD generation.
It distills heterogeneous expert knowledge, including CATIA operation recordings, macro logs, drawing notes, engineering descriptions, and simple CAD instructions, into \textbf{CAD-IR}, an executable intermediate representation with explicit parameters, ordered operations, tool bindings, outputs, dependencies, and verification rules.
By further abstracting CAD-IR into reusable and parameterized CAD skills, the mechanism enables expert modeling procedures to be adapted to new variant requests within the same component family.

\item We develop \algname, a skill-guided industrial CAD agent with multi-backend execution support.
\algname retrieves expert-derived skills, instantiates and revises CAD procedures at the IR level, and dispatches executable operations to different CAD backends.
In this work, we implement a dedicated \textbf{CATIA-MCP} backend for CATIA-native industrial B-Rep modeling and incorporate a lightweight multi-view vision feedback loop to refine CAD-IR during iterative generation.

\item We empirically validate \algname on both public text-to-CAD tasks and complex industrial automotive components. 
On the Text2CAD dataset, CAD-IR improves generation performance under intermediate-level prompts, showing its effectiveness in bridging vague textual descriptions and executable CAD procedures. 
In addition, we record expert macro demonstrations for four complex automotive parts, convert them into CAD-IR and reusable skills, and demonstrate that \algname can generate industrial-grade CAD models through CATIA-MCP.

\end{itemize}

\section{Related work}
\subsection{Text-Guided CAD Generation and CAD Agents}

Learning-based CAD generation has progressively moved from final-shape prediction toward structured procedural modeling.
Early works represent CAD designs as sketches, constraints, graphs, or command sequences, making CAD history a learnable target rather than treating the final geometry as an isolated 3D shape.
SketchGraphs models relational structures among sketch entities and constraints~\cite{seff2020sketchgraphs}, while Computer-Aided Design as Language serializes CAD sketches into structured tokens and applies language modeling techniques to sketch generation~\cite{ganin2021computer}.
Fusion 360 Gallery releases large-scale human design sequences and enables data-driven studies of programmatic CAD construction~\cite{willis2021fusion}.
DeepCAD further represents 3D CAD models as sketch-and-extrude command sequences and learns a Transformer-based generative model~\cite{wu2021deepcad}.
SkexGen improves controllable CAD sequence generation by disentangling topology, geometry, and extrusion information~\cite{xu2022skexgen}.
These methods establish procedural CAD modeling as a promising direction, but they are mainly designed for benchmark-scale parametric programs and relatively regular sketch-extrude geometries.

Recent text-to-CAD methods introduce natural language as a more intuitive interface for CAD generation.
Text2CAD maps textual descriptions to parametric CAD command sequences under different description granularities~\cite{khan2024text2cad}.
CAD Translator studies translation across CAD-related modalities and representations~\cite{li2024cad}.
CAD-GPT and CAD-LLaMA explore large language models for generating CAD commands or programs from text~\cite{wang2025cad,li2025cad}.
CADFusion further incorporates multimodal conditioning to improve controllability~\cite{wang2025text}.
Beyond direct sequence or program generation, recent CAD-agent systems connect language models with executable CAD tools.
CAD-Assistant uses multimodal language models and CAD tool execution for interactive modeling~\cite{mallis2025cad}, while FutureCAD studies B-Rep-aware CAD generation from textual queries~\cite{li2026towards}.
These works show that language models can connect design intent with executable CAD operations, and that tool execution and feedback are useful for improving controllability.

Despite these advances, existing text-guided CAD systems are still far from industrial CAD workflows.
Most methods generate command sequences or CAD programs from text alone and are evaluated on public benchmarks with limited geometric and procedural complexity.
They do not explicitly exploit expert modeling records that naturally exist in industrial practice, such as macro logs, feature trees, operation histories, parameter tables, and reusable modeling templates.
As a result, they lack a mechanism to transfer human procedural knowledge into new CAD tasks.
In contrast, \algname focuses on expert-grounded industrial CAD generation.
It converts natural-language descriptions, simple CAD instructions, and expert macro demonstrations into executable CAD-IR, distills expert procedures into reusable parameterized skills, and dispatches the resulting procedures through backend execution layers such as CATIA-MCP.
This allows \algname to reuse human modeling knowledge and target complex production-oriented B-Rep components rather than only generating isolated benchmark CAD sequences.

\section{Method}
\label{sec:method}

\subsection{Overview}

\algname is an IR-centered industrial CAD agent that converts expert procedural knowledge into reusable CAD skills and uses these skills to generate new CAD variants from user requests.
The framework consists of two stages: \emph{expert-to-skill acquisition} and \emph{skill-guided variant CAD generation}.
The first stage is an offline knowledge construction process.
Given expert CATIA operation recordings, macro logs, drawing notes, feature trees, and parameter tables, \algname extracts the modeling procedure of a component family and distills it into a structured part skill.
Each skill contains four key elements: a basic part description, an expert operation order, important parameter schemas and ranges, and a template CAD-IR.
The second stage is an online generation process.
Given a user-provided \emph{variant request}, \algname retrieves a relevant skill, instantiates its template CAD-IR, executes the generated procedure through a CAD backend, and revises the IR according to execution and visual feedback.

The key design choice is that \algname does not directly generate backend-specific scripts.
Instead, it represents CAD construction as \textbf{CAD-IR}, an executable intermediate representation that explicitly records the ordered MCP tool calls, required parameters, operation dependencies, generated entities, and verification rules.
This design provides two practical advantages.
First, CAD-IR serves as a common procedural layer between expert demonstrations, language instructions, and backend execution.
Second, CAD-IR allows the agent to revise a CAD procedure at the operation level rather than rewriting a long backend script from scratch.
As a result, skill-guided IR generation provides stronger procedural guidance, reduces the search space of the agent, and improves robustness for ambiguous or underspecified industrial CAD requests.

\subsection{Expert-to-Skill Acquisition with CAD-IR}

The goal of expert-to-skill acquisition is to transform expert modeling experience from one-off CAD records into reusable procedural knowledge.
In industrial CAD workflows, senior engineers' CATIA operations can be recorded as macro logs, which directly preserve the executable modeling process behind a part.
These recordings capture key procedural decisions such as sketch construction, reference creation, feature generation, trimming or splitting, parameter assignment, and export.
Therefore, parsing expert CATIA recordings into ordered operations is the core step for distilling human CAD knowledge into reusable skills.

For each component family, we collect expert experience from CATIA macro recordings, drawing notes, and descriptions.
These heterogeneous records are distilled into a part skill:
\begin{equation}
s = {\mathcal{D}_s, \mathcal{O}_s, \mathcal{P}_s, \mathcal{Z}_s},
\end{equation}
where $\mathcal{D}_s$ denotes the basic information of the component family (including descriptions, symmetry, smoothness, and boundaries) , $\mathcal{O}_s$ denotes the expert operation order, $\mathcal{P}_s$ denotes the parameter schema and valid ranges, and $\mathcal{Z}_s$ denotes the template CAD-IR.

\paragraph{Part description.}
The part description $\mathcal{D}_s$ summarizes the target component family, its functional role, and its major geometric characteristics.
For example, an automotive reinforcement plate skill may describe the base plate, mounting holes, bending regions, stiffening ribs, boundary flanges, and symmetry constraints that are commonly shared by different variants of the same component.

\paragraph{Expert operation order.}
The expert operation order $\mathcal{O}_s$ captures the modeling sequence extracted from CATIA macro recordings.
A recorded macro is first parsed into low-level operations, and these operations are then grouped into semantically meaningful stages, such as creating the base sketch, constructing guide curves, sweeping a surface, trimming or splitting geometry, adding mounting holes, and finalizing visibility or export.
This operation order is not treated as a fixed replay script.
Instead, it provides a reusable procedural prior that describes how experts usually construct this component family.
This prior is critical for complex parts because many modeling decisions, such as which reference to create first and which feature should be delayed until later, are difficult to infer from text alone.

\paragraph{Parameter schema.}
The parameter schema $\mathcal{P}_s$ defines the key editable parameters of the component family.
Typical parameters include width, length, depth, radius, hole number, hole diameter, arc height, flange radius, extrusion length, and sweep curvature.
For each parameter, the skill records its semantic meaning, default value, and valid range when available.
These parameters allow the same expert-derived skill to be adapted to new variant requests without rebuilding the full CAD procedure from scratch.

\paragraph{Template CAD-IR.}
The template CAD-IR $\mathcal{Z}_s$ is the executable procedural template distilled from expert operations.
We define CAD-IR as:
\begin{equation}
\mathcal{Z} = {\mathcal{P}, \mathcal{T}, \mathcal{O}, \mathcal{E}, \mathcal{V}},
\end{equation}
where $\mathcal{P}$ is the set of global part parameters, $\mathcal{T}$ is the set of required backend tools, $\mathcal{O}$ is an ordered list of CAD operations, $\mathcal{E}$ records generated entities and dependency relations, and $\mathcal{V}$ contains verification rules.
Each operation in CAD-IR is represented as:
\begin{equation}
o_i = {\mathrm{id}, \mathrm{intent}, \mathrm{mcp\_calls}, \mathrm{arguments}, \mathrm{outputs}, \mathrm{dependson}, \mathrm{verify}}.
\end{equation}
Here, \texttt{intent} describes the modeling purpose of the operation, \texttt{mcp\_calls} specifies the backend tool chain to be executed, \texttt{arguments} stores concrete or symbolic parameters, \texttt{outputs} records newly created entities, \texttt{dependson} defines operation dependencies, and \texttt{verify} specifies execution checks.

In our implementation, this acquisition stage is performed by a code-oriented agent, Codex-xHigh, which parses CATIA macro recordings and experts' descriptions and drawing notes into normalized operations, abstracts repeated parameter patterns, and writes the resulting skill and template CAD-IR.
The generated template is then validated by replaying the corresponding MCP tool chain.
This process converts expert demonstrations from one-off macro scripts into reusable, parameterized, and executable CAD skills.

\subsection{Skill-Guided Variant CAD Generation}

\begin{figure}
\centering
\includegraphics[width=\linewidth]{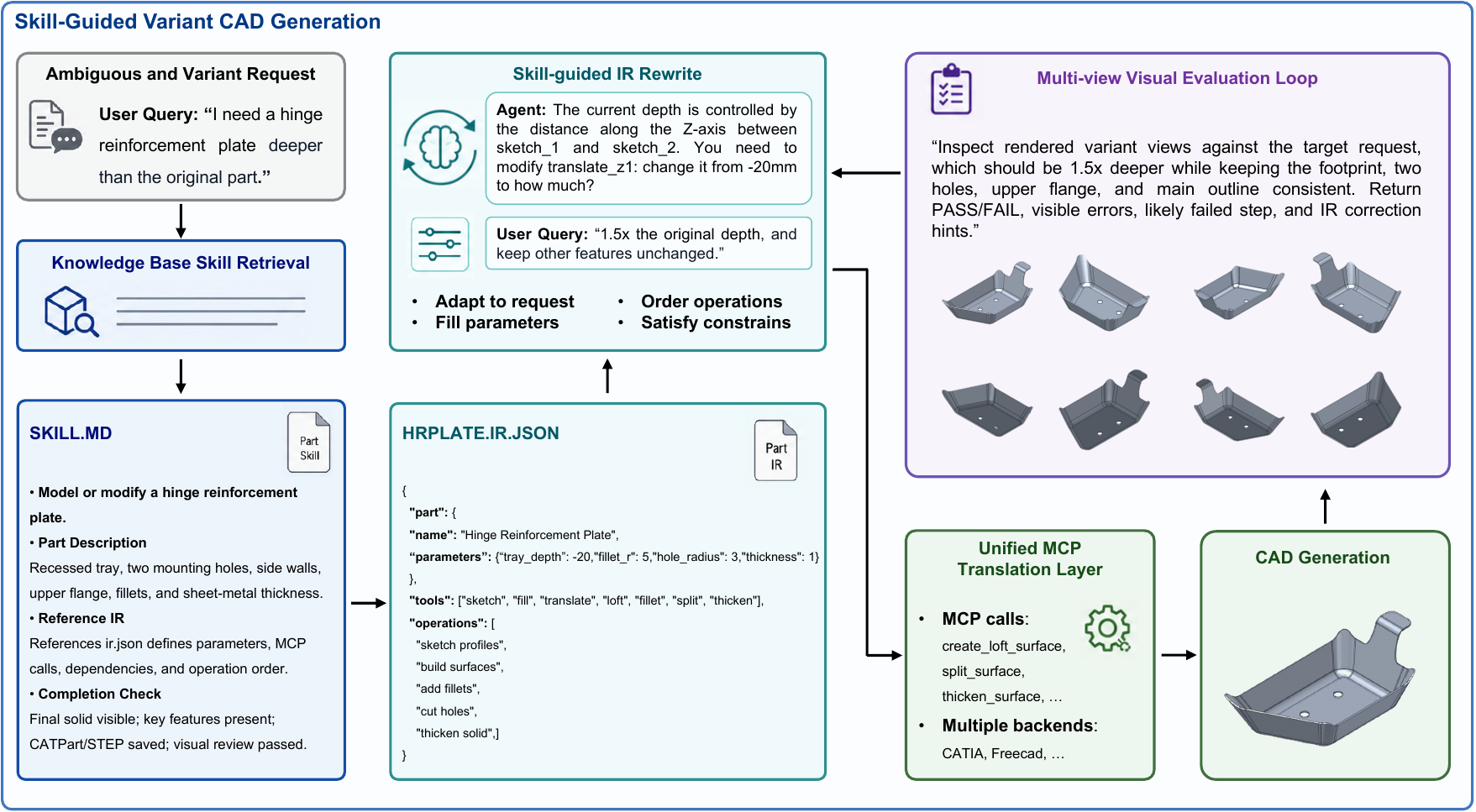}
\caption{
Overview of the skill-guided variant CAD generation framework.
Given an ambiguous or variant user request, \algname retrieves an expert-derived part skill, instantiates and rewrites a structured CAD-IR, translates the IR into backend-specific MCP calls for CAD execution, and uses multi-view visual feedback to iteratively refine the IR until the generated CAD variant satisfies the target request.
}
\label{fig:framework}
\end{figure}

Given a user-provided variant request $\mathcal{R}$, \algname generates a new CAD model by adapting an expert-derived skill rather than constructing the entire procedure from scratch.
A variant request refers to a new requirement within the same component family, such as changing the depth, width, curvature, hole layout, local reinforcement structure, or boundary shape of an existing part.
The online generation stage contains skill matching, CAD-IR instantiation, backend execution, and vision-guided IR rewriting, as illustrated in Figure~\ref{fig:framework}.

\subsubsection{Skill Matching and CAD-IR Instantiation}

Given an ambiguous or variant user request $\mathcal{R}$, the key challenge is to convert incomplete design intent into a complete and executable CAD procedure.
A request such as ``draw a deeper hinge reinforcement plate with mounting holes'' usually specifies only high-level changes, but leaves many modeling details unspecified, including the base construction sequence, reference entities, parameter dependencies, operation order, and verification rules.
\algname addresses this problem by grounding IR generation in expert-derived skills.

The agent first retrieves the most relevant skill from the skill knowledge base $\mathcal{K}$:
\begin{equation}
s^\star = \mathrm{Retrieve}(\mathcal{R}, \mathcal{K}).
\end{equation}
The retrieved skill provides a component-level modeling prior, including the expected part structure, expert operation order, parameter schema, and template CAD-IR.
Conditioned on this skill, the agent interprets the variant request as a set of structured IR edits rather than generating a CAD program from scratch.
Specifically, it identifies which parameters should be changed, which local operations should be inserted or modified, which constraints should be preserved, and which expert operation order should remain unchanged.

The initial CAD-IR for the new variant is obtained by instantiating the template IR:
\begin{equation}
\mathcal{Z}*0 = \mathrm{Instantiate}(\mathcal{Z}*{s^\star}, \mathcal{R}),
\end{equation}
where $\mathcal{Z}_{s^\star}$ is the template CAD-IR stored in the retrieved skill.
This instantiation process fills missing parameters, adapts geometric dimensions, updates operation arguments, orders the required MCP tool calls, and preserves dependencies and verification rules.
In this way, the skill acts as a procedural scaffold: it turns an underspecified request into a complete CAD-IR with executable operations, valid references, and checkable outputs.

In our implementation, skill matching, CAD-IR instantiation, and subsequent IR rewriting are performed by Mimo-v2.5-Pro.
The model is prompted with the variant request, the retrieved skill, the parameter schema, the template CAD-IR, and any available execution or visual feedback.
It outputs either a complete instantiated CAD-IR or a structured IR patch that modifies selected parameters, operations, dependencies, or verification rules.
This skill-guided formulation improves generation efficiency and robustness because the agent only needs to adapt an expert-derived procedure to the requested variant, rather than rediscovering the full industrial CAD workflow from an ambiguous prompt.

\subsubsection{Backend Execution with CAD-MCP}

After IR instantiation, \algname dispatches the generated CAD-IR to a backend execution layer.
The execution layer reads the ordered operations in $\mathcal{Z}_t$ and translates each operation into backend-specific MCP tool calls:
\begin{equation}
e_t, \mathcal{M}*t = \mathrm{Exec}*{\mathrm{MCP}}(\mathcal{Z}_t),
\end{equation}
where $\mathcal{M}_t$ is the generated CAD model and $e_t$ contains execution status, error messages, produced entities, and diagnostic information.

In this work, we implement a CATIA-MCP backend for CATIA-native industrial CAD construction.
The backend supports the tool chain required by our evaluated automotive components, including part creation, hybrid body construction, sketch creation, profile construction, sweep, extrusion, datum creation, split, visibility control, saving, and export.
CATIA-MCP is not designed as a standalone algorithmic contribution.
Its role is to provide a reliable industrial execution interface for CAD-IR, enabling \algname to generate editable B-Rep models in a production CAD environment.

\subsubsection{Vision Feedback and IR Rewrite}

Execution feedback alone cannot guarantee that the generated model satisfies the user's visual and geometric intent.
Therefore, \algname uses a lightweight vision feedback loop to refine CAD-IR.
After backend execution, the system exports canonical multi-view renderings of the current model:
\begin{equation}
\mathcal{I}_t = {I_t^{front}, I_t^{back}, I_t^{left}, I_t^{right}, I_t^{top}, I_t^{bottom}, I_t^{iso1}, I_t^{iso2}}.
\end{equation}
These eight views are merged into a visual board and provided to Mimo-v2.5-Pro together with the variant request, the retrieved skill, and a summary of the current CAD-IR.
The model returns visual feedback $v_t$ describing missing structures, incorrect proportions, violated visual constraints, or failed modeling intentions.

The agent then revises the CAD-IR rather than directly editing backend-specific scripts:
\begin{equation}
\mathcal{Z}_{t+1} = \mathrm{Rewrite}(\mathcal{Z}_t, \mathcal{R}, e_t, v_t).
\end{equation}
The rewrite may update parameter values, modify operation arguments, insert missing operations, remove invalid steps, or adjust dependencies while preserving the overall expert-derived operation structure.
The updated IR is executed again through the backend.
This loop continues until the generated model passes the visual evaluation or reaches the maximum refinement budget.

\section{Experiments}
\label{sec}

We evaluate \algname from two complementary perspectives.
First, we use the public Text2CAD benchmark to test whether CAD-IR can improve generation from simple and intermediate-level text prompts without using any expert knowledge base.
Second, we evaluate \algname on industrial automotive components, where expert CATIA macro recordings are converted into reusable skills and used to generate new design variants requested by CAD experts.
In both settings, the final outputs of \algname are executed through CATIA-MCP and exported as editable B-Rep CAD models.

\subsection{Evaluation on the Public Text2CAD Benchmark}

\paragraph{Setting.}
We first evaluate \algname on the public Text2CAD dataset.
This experiment is designed to isolate the benefit of CAD-IR itself.
Therefore, we do not use any expert macro logs, industrial skill library, or component-specific knowledge base for this benchmark.
Given an intermediate-level text prompt, \algname directly generates CAD-IR, executes the resulting procedure through CATIA-MCP, and exports the generated model as a CATIA-native B-Rep model.
We randomly sample $100$ test cases from the intermediate-prompt split for evaluation.

This setting is intentionally challenging because intermediate prompts are often ambiguous: they describe the target geometry at a coarse semantic level but omit many low-level construction details, such as exact operation order, reference selection, and parameter dependencies.
The goal is to test whether CAD-IR can serve as a useful procedural scaffold that improves generation quality even without expert-derived skills.

\paragraph{Baselines.}
We compare \algname with representative text-to-CAD baselines and internal variants:
\begin{itemize}
\item \textbf{Text2CAD}: the original text-to-CAD baseline that maps textual descriptions to CAD command sequences~\cite{khan2024text2cad}.
\item \textbf{CAD-Coder}: a code-generation baseline for producing executable CAD procedures from text~\cite{guan2026cad}.
\item \textbf{\algname w/o IR}: an ablated version that directly generates CAD operations from the intermediate prompt without using CAD-IR as an explicit intermediate representation.
\item \textbf{\algname w/o IR (expert)}: an upper-reference setting where the model receives expert-level descriptions instead of intermediate prompts, but still does not use CAD-IR.
\item \textbf{\algname}: our full public-benchmark setting, where the model receives only intermediate prompts and directly generates executable CAD-IR.
\end{itemize}

\paragraph{Metrics.}
We evaluate both executability and geometric fidelity.
For executability, we report the number of valid outputs among $100$ generated samples.
For geometric fidelity, we follow common text-to-CAD evaluation protocols and report Chamfer Distance $\mathrm{CD}\times 10^3$~\cite{fan2017point} and solid IoU at resolution $32^3$.
Lower CD indicates better surface-level geometry matching, while higher solid IoU indicates better volumetric consistency.
For \algname, all metrics are computed after executing the generated CAD-IR in CATIA and exporting the resulting B-Rep model.

\begin{table}[t]
\centering
\small
\caption{Results on $100$ randomly sampled intermediate-level prompts from the Text2CAD benchmark.
\algname directly generates CAD-IR from intermediate prompts without using an expert skill library.
All \algname outputs are executed through CATIA-MCP and exported as editable B-Rep CAD models.}
\label{tab:text2cad_results}
\resizebox{\linewidth}{!}{
\begin{tabular}{lccccc}
\toprule
Method & Prompt Level & Valid / Total & CD Mean $\downarrow$ & CD Median $\downarrow$ & Solid IoU Mean $\uparrow$ \\
\midrule
\algname w/o IR (expert) & Expert & $100/100$ & $10.38$ & $0.96$ & $0.662$ \\
\midrule
Text2CAD & Intermediate & $96/100$ & $29.80$ & $10.19$ & $0.446$ \\
CAD-Coder & Intermediate & $100/100$ & $14.16$ & $3.47$ & $0.585$ \\
\algname w/o IR & Intermediate & $100/100$ & $14.83$ & $2.45$ & $0.614$ \\
\algname & Intermediate & $100/100$ & $\mathbf{9.88}$ & $\mathbf{2.12}$ & $\mathbf{0.646}$ \\
\bottomrule
\end{tabular}
}
\label{Table:text2cad}
\end{table}

\paragraph{Results and analysis.}
As shown in Table~\ref{Table:text2cad}, CAD-IR substantially improves intermediate-level text-to-CAD generation.
Compared with \algname w/o IR under the same intermediate-prompt setting, \algname reduces CD mean from $14.83$ to $9.88$ and improves solid IoU mean from $0.614$ to $0.646$.
This indicates that CAD-IR provides useful procedural structure even when no expert skills are available.
Compared with external baselines, \algname achieves consistent gains under the intermediate-prompt setting.
It improves CD Mean by $66.8\%$ over Text2CAD and $30.2\%$ over CAD-Coder, while improving Solid IoU Mean from $0.446$ and $0.585$ to $0.646$, respectively.
This indicates that CAD-IR provides a stronger procedural scaffold for translating ambiguous intermediate prompts into executable CAD procedures.

More importantly, \algname with intermediate prompts reaches a performance level close to the expert-description setting.
Although \algname only receives intermediate prompts, its CD mean is even lower than that of \algname w/o IR (expert), and its solid IoU mean remains competitive.
This suggests that CAD-IR can partially bridge the gap between vague textual descriptions and expert-level procedural CAD construction.
The result supports our central claim: representing CAD generation as an explicit executable IR can improve both robustness and geometry quality, even before introducing expert-derived skills.

\begin{figure}
\centering
\includegraphics[width=\linewidth]{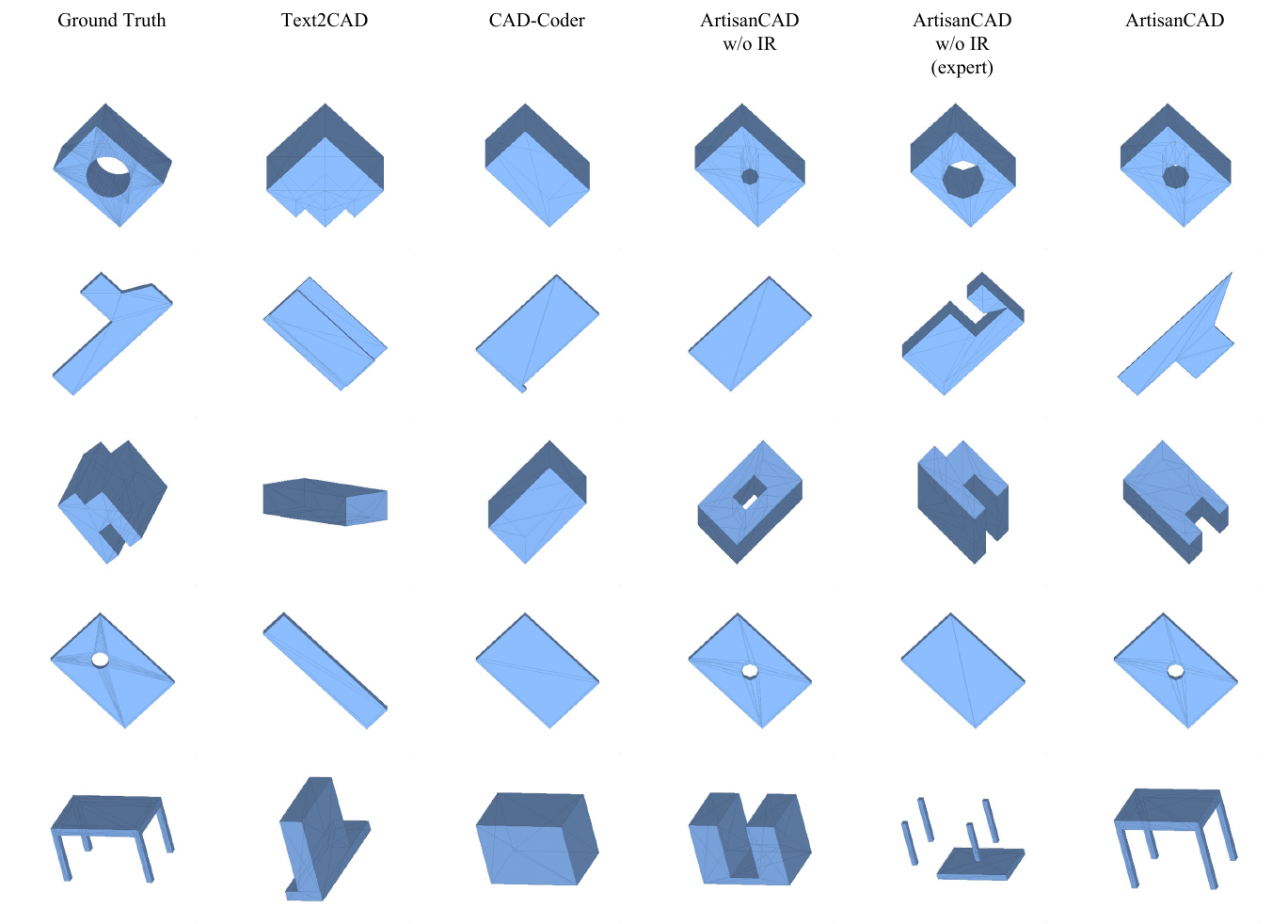}
\caption{Qualitative comparison on representative Text2CAD samples.
Each row shows one target shape, and each column compares the ground truth with Text2CAD, CAD-Coder, \algname without CAD-IR, \algname without CAD-IR using expert descriptions, and the full \algname.
The visual results show that CAD-IR helps preserve important part-level structures and local details under intermediate prompts.}
\label{fig:sample_vision}
\end{figure}

\paragraph{Qualitative visualization.}
Fig.~\ref{fig:sample_vision} provides representative visual examples that complement the quantitative results.
Compared with the two external baselines, \algname produces more faithful CAD structures under the intermediate-prompt setting.
Unlike expert prompts, intermediate prompts do not provide a detailed modeling procedure, reference construction strategy, feature order, or explicit operation decomposition.
Without such step-by-step procedural guidance, Text2CAD and CAD-Coder often fail to infer a complete CAD construction process: they recover only a coarse bounding shape, while missing functional details such as through holes, local cutouts, branching surfaces, and thin supporting members.
For example, in the table-like sample in the last row, Text2CAD and CAD-Coder produce bulky or collapsed solids rather than a recognizable table structure.
By contrast, \algname reconstructs a flat tabletop with four slender legs and preserves the spatial relationship between the supports and the top plate.

The benefit of CAD-IR is further illustrated by the comparison between \algname and its non-IR variant.
Because CAD-IR explicitly represents parameters, ordered operations, dependencies, and verification rules, it helps the agent infer the missing procedural information that is not stated in intermediate prompts.
This effect is visible in the second, third, and fifth rows.
In the second row, \algname w/o IR degenerates to a simple prism-like body, whereas the full \algname recovers the branched profile and asymmetric extension visible in the target.
In the third row, CAD-IR helps preserve the local slot/cutout structure and the overall part orientation, while the non-IR variant produces a less faithful block-like shape.
In the fifth row, the non-IR result separates the object into coarse disconnected blocks, but the full \algname reconstructs the tabletop and supporting legs as a coherent CAD part.
These examples suggest that CAD-IR does not merely improve global shape similarity; it provides an operation-level scaffold that helps the agent preserve functional details and compose them into editable CAD structures.

\subsection{Industrial Evaluation on Automotive CAD Components}

\paragraph{Setting.}
The second experiment evaluates whether expert-derived skills can support industrial CAD generation.
We select four relatively complex automotive components from real vehicle CAD workflows.
For each component, expert engineers operate CATIA to construct the part, and we record the corresponding macro logs and engineering notes.
Each macro group is then converted into a part-specific skill through the expert-to-skill acquisition pipeline described in Sec.~\ref{sec}.
In total, we obtain four expert-derived skills, one for each component family.

Each skill contains a part description, an expert operation order, a parameter schema, and a template CAD-IR.
The template CAD-IR stores the ordered MCP tool chain, required parameters, operation dependencies, generated entities, and verification rules.
During evaluation, CAD experts provide additional variant requests for each component family.
These requests modify dimensions, depths, hole layouts, local reinforcements, curvature, or boundary shapes while preserving the high-level component identity.
Given a variant request, \algname retrieves the corresponding skill, instantiates the template CAD-IR, executes it through CATIA-MCP, and refines the IR using the vision feedback loop. In the ablation study, we try to directly generate four complex automotive components without CAD-IR.

\begin{figure}[t]
\centering
\includegraphics[width=\linewidth]{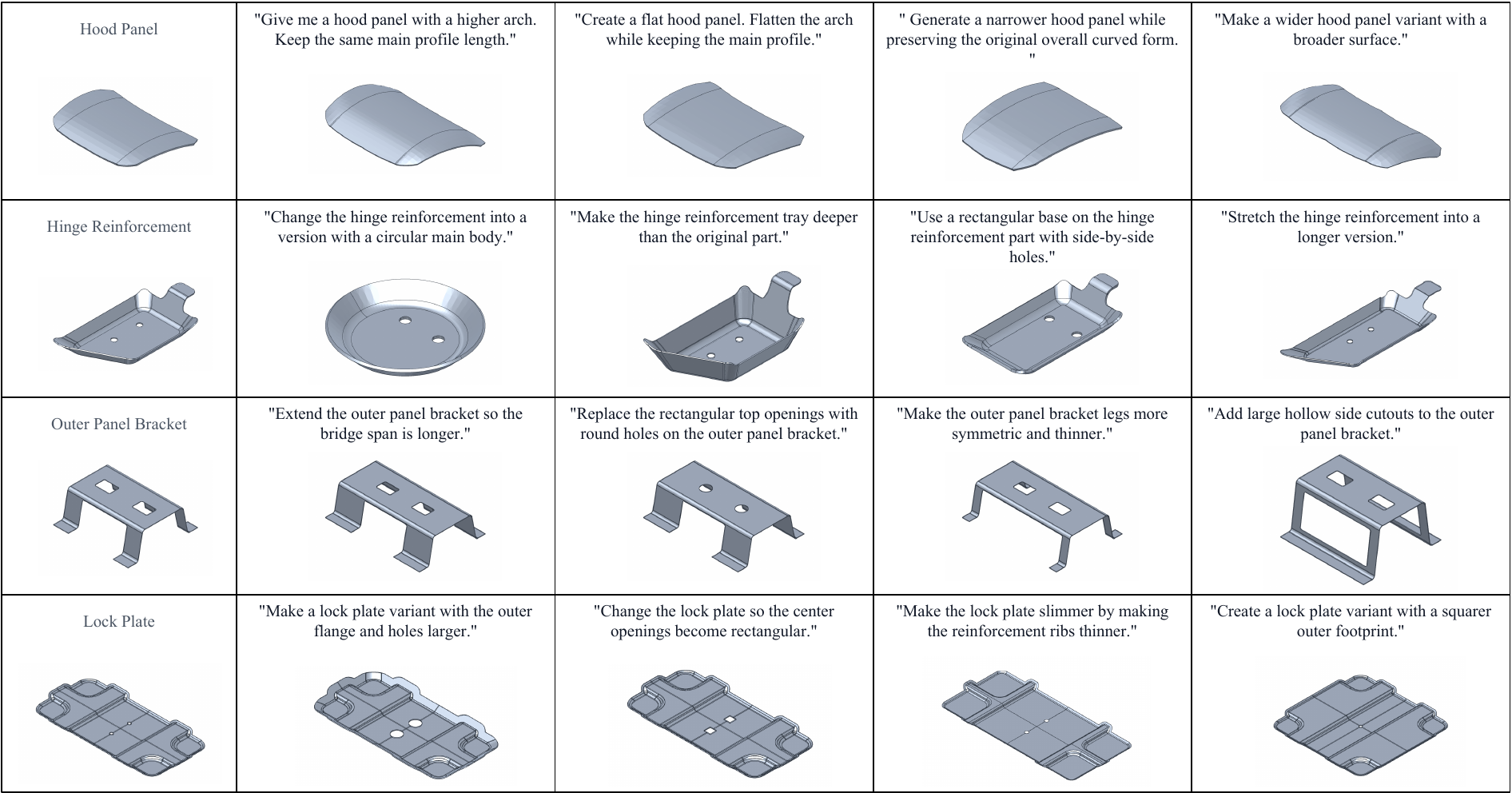}
\caption{Industrial variant generation with expert-derived skills.
Each row corresponds to one automotive component family, where the first column shows the original expert-modeled part and the remaining columns show new variants generated from expert-provided variant requests.
\algname adapts the corresponding skill by modifying parameters and local operations in CAD-IR, and executes the resulting procedure through CATIA-MCP to produce editable B-Rep CAD models.}
\label{fig:industrial_variants}
\end{figure}

\paragraph{Results.}
Fig.~\ref{fig:industrial_variants} shows qualitative results on four automotive component families.
Given expert-provided variant requests, such as changing the arch height, width, depth, hole layout, base shape, cutout structure, flange size, or reinforcement ribs, \algname successfully adapts the corresponding expert-derived skill and generates executable CAD-IR for each new variant.
The generated IR preserves the expert operation structure while modifying only the required parameters and local operations, and can be executed through CATIA-MCP to produce editable CATIA-native B-Rep models.
In contrast, without skill-guided CAD-IR, the agent fails on these industrial components because it cannot reliably infer the long-horizon operation order, valid references, and parameter dependencies from short variant requests alone.
This demonstrates that CAD-IR is essential for transferring expert modeling knowledge to new production-oriented CAD variants.

\paragraph{Analysis.}
This experiment demonstrates the practical value of expert-to-skill acquisition.
For complex industrial components, the main difficulty is not only selecting individual CAD commands, but also knowing the correct procedural structure: which references to build, how to order dependent operations, which parameters should remain editable, and how to preserve valid feature dependencies.
Expert macro recordings provide this missing procedural knowledge.
By distilling them into CAD-IR and reusable skills, \algname can reuse expert modeling experience to generate new component variants.

The contrast between the two settings also highlights the role of skills in industrial CAD generation.
A general-purpose CAD agent with access to CATIA tools is insufficient when the task requires long operation chains and domain-specific modeling conventions.
Expert-derived skills make the problem tractable by providing a structured template that the agent can adapt rather than rediscover.
Together with CATIA-MCP execution and vision-based IR rewriting, this enables \algname to move from public benchmark-style text-to-CAD generation toward production-oriented CAD automation.

\section{Conclusion}

We presented \algname, a skill-guided industrial CAD agent that uses CAD-IR as an executable intermediate representation for expert-grounded CAD generation. 
CAD-IR serves both as a carrier for distilling expert CATIA recordings and engineering notes into reusable parameterized skills, and as a procedural scaffold for converting ambiguous or intermediate-level prompts into complete executable CAD operations. 
Through skill retrieval, IR instantiation, CATIA-MCP execution, and multi-view vision feedback, \algname generates editable, production-ready B-Rep models for new variant requests. 
Experiments on the Text2CAD benchmark show that CAD-IR improves intermediate-prompt generation, while industrial experiments on four complex automotive components demonstrate that expert-derived skills are essential for scaling CAD agents toward practical production workflows.

\section*{Acknowledgments}

Acknowledgments will be added in the final version.

\bibliographystyle{unsrt}
\bibliography{references}

\end{document}